\pdfoutput=1

\documentclass[11pt]{article}
\usepackage{authblk}
\usepackage{natbib}

\usepackage[]{ACL2023}

\usepackage{times}
\usepackage{latexsym}
\usepackage{tabularray}
\usepackage{array}
\usepackage{graphicx}
\usepackage{multirow}
\usepackage{comment}
\usepackage{longtable}
\usepackage{booktabs}
\usepackage[T1]{fontenc}
\usepackage[utf8]{inputenc}

\newcounter{notecounter}

\newcommand{\enoteson}{\long\gdef\enote##1##2{{
			\stepcounter{notecounter}
			{\large\textbf{ \hspace{1cm}\arabic{notecounter} $<<<$ ##1: ##2 $>>>$\hspace{1cm}}}}}}
\enoteson

\usepackage[utf8]{inputenc}

\usepackage{microtype}
\usepackage{xspace}

\usepackage{inconsolata}

\def\dataset{TurkishMMLU\xspace}
\def\datasetsmall{TurkishMMLU\textsubscript{sub}\xspace}

\title{TurkishMMLU: Measuring Massive Multitask Language Understanding in Turkish}

\author[1]{Arda Yüksel}
\author[2,3,4]{Abdullatif Köksal}
\author[2,3]{Lütfi Kerem Senel}
\author[4]{\\Anna Korhonen}
\author[2,3]{Hinrich Sch\"utze}

\affil[1]{Technical University of Munich}
\affil[2]{Center for Information and Language Processing, LMU Munich}
\affil[3]{Munich Center for Machine Learning}
\affil[4]{Language Technology Lab, University of Cambridge \protect\\
\texttt{arda.yueksel@tum.de}, \texttt{akoksal@cis.lmu.de}}

\begin{document}
\maketitle
\begin{abstract}
Multiple choice question answering tasks evaluate the reasoning, comprehension, and mathematical abilities of Large Language Models (LLMs). While existing benchmarks employ automatic translation for multilingual evaluation, this approach is error-prone and potentially introduces culturally biased questions, especially in social sciences. We introduce the first multitask, multiple-choice Turkish QA benchmark, \dataset, to evaluate LLMs' understanding of the Turkish language. \dataset includes over 10,000 questions, covering 9 different subjects from Turkish high-school education curricula. These questions are written by curriculum experts, suitable for the high-school curricula in Turkey, covering subjects ranging from natural sciences and math questions to more culturally representative topics such as Turkish Literature and the history of the Turkish Republic. We evaluate over 20 LLMs, including multilingual open-source (e.g., Gemma, Llama, MT5), closed-source (GPT 4o, Claude, Gemini), and Turkish-adapted (e.g., Trendyol) models. We provide an extensive evaluation, including zero-shot and few-shot evaluation of LLMs, chain-of-thought reasoning, and question difficulty analysis along with model performance. We provide an in-depth analysis of the Turkish capabilities and limitations of current LLMs to provide insights for future LLMs for the Turkish language. We publicly release our code for the dataset and evaluation: \url{https://github.com/ArdaYueksel/TurkishMMLU}.
\end{abstract}

\section{Introduction}

\begin{figure}[t!]
    \centering
    \includegraphics[width=\columnwidth]{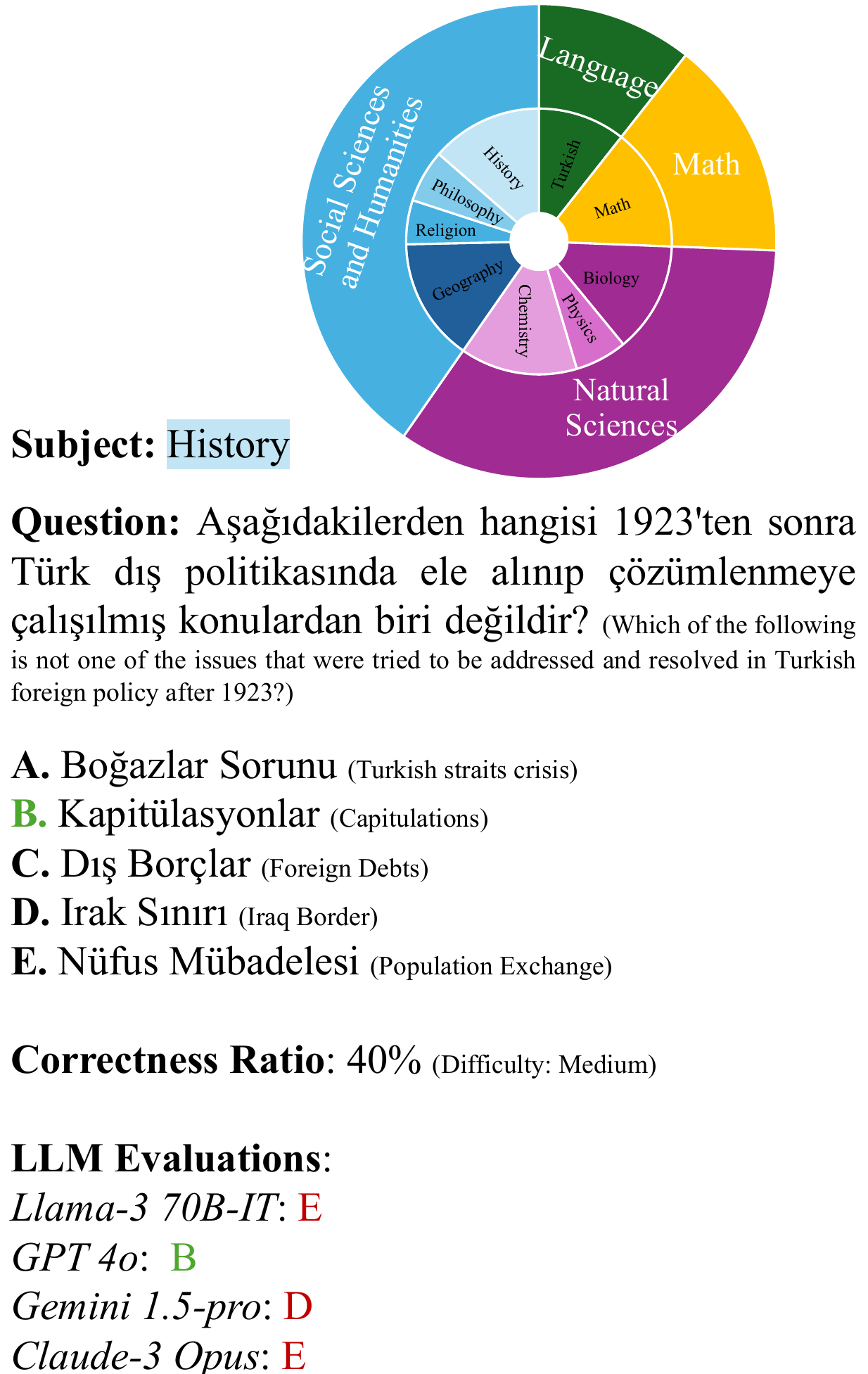}
    \caption{The chart displays the subject distribution of \dataset. An example from our dataset shows recent multilingual LLMs struggling with a question about Turkish history.}
    \label{fig:coverage_pie}
\end{figure}

Benchmarking plays an important role in understanding and measuring the capabilities of language models. Recent multitask multiple-choice question answering (QA) benchmarks like MMLU \citep{hendrycks2021mmlu} cover a wide range of use cases for language models, making them highly popular as one of the main evaluation benchmarks in recent LLMs such as GPT 4 \citep{openai2024gpt4} and Gemini \citep{geminiteam2024gemini}. For the multilingual adaptation of the MMLU benchmark, recent works \citep{lai-etal-2023-okapi} have focused on automatic translations. However, automatic translations are often prone to errors and may fail to capture the linguistic and cultural nuances of the target language. Consequently, there have been manual efforts to create multitask multiple-choice benchmarks in various languages, including Arabic \citep[ArabicMMLU,][]{koto2024arabicmmlu}, Korean \citep[KMMLU,][]{son2024kmmlu}, and Chinese \citep[CMMLU,][]{li2023cmmlu}.

In our work, we introduce \dataset, the first multitask multiple-choice QA benchmark specifically designed for the Turkish language. Our dataset includes 10,032 multiple-choice questions, each with five options, spanning nine subjects categorized into four groups: \textit{Natural Sciences}, \textit{Mathematics}, \textit{Turkish Language and Literature}, and \textit{Social Sciences and Humanities}. These questions are sourced from a high-quality online learning platform created by the Turkish Ministry of Education, which aims to support high school students in preparing for the university entrance exam. A unique feature of \dataset is the correctness ratio, which reflects the actual performance of students on these questions, offering a more accurate measure of question difficulty. We illustrate the distribution of subjects and an example from \dataset in Figure \ref{fig:coverage_pie}.

After introducing this dataset for benchmarking in Turkish,
we evaluate a wide range of current language models, more
than 40, including multilingual autoregressive LLMs, both
open models like Gemma \citep{gemmateam2024gemma}, Llama-3
and Aya-23 \citep{aryabumi2024aya} and closed-source models
such as GPT 4o, Claude and Gemini. In addition, we also
cover multilingual encoder-decoder models such as MT5, MT0,
Aya and Turkish-adapted LLMs such as Trendyol-LLM, a LoRA adaptation of multilingual LLMs. We also cover many different setups including zero-shot, few-shot, and chain-of-thought \citep{wei2022chain}. We further provide analysis of LLMs based on subjects and difficulty. Our additional analysis provides insights for the design of future LLMs for Turkish and beyond. We publicly release our code for the dataset and evaluation: \url{https://github.com/ArdaYueksel/TurkishMMLU}.

Our contributions are as follows:
\begin{enumerate}
    \item We introduce the first large-scale multitask multiple-choice benchmark for Turkish, consisting of 10,032 questions across nine subjects.
    \item We evaluate a wide range of LLMs, varying in size from 60M to 141B, including both open and closed-source models, and provide a comprehensive leaderboard featuring over 40 models.
    \item We conduct an in-depth analysis of LLM performance in chain-of-thought setups and based on question difficulty.
\end{enumerate}

\section{Related Work}\label{sec:related}

\textbf{LLM Benchmarking:} 
Benchmarks are crucial for understanding the capabilities of NLP models, identifying their weaknesses and facilitating the development of more capable models.
Historically, most NLP benchmarks focused on linguistic tasks \citep{wang-etal-2018-glue,wang2019superglue,rajpurkar-etal-2016-squad} and followed the paradigm of supervised fine-tuning of a model on a training set and evaluation on an unseen test set.
However, with the advent of powerful LLMs, this type of evaluation became obsolete as these models showed impressive zero-shot and few-shot learning skills, even for higher level tasks closer to real world applications. 
To evaluate the emerging capabilities of the LLMs, new benchmarks are proposed that focus on more advanced capabilities such as common sense reasoning \citep{10.5555/3031843.3031909}, multi-hop reasoning \citep{yang-etal-2018-hotpotqa}, programming \citep{chen2021evaluating} and multi-turn conversations. 
Additionally, some studies aimed at evaluating these capabilities through extensive datasets that cover a broad range of knowledge-based topics \citep{srivastava2023beyond}. 
One prominent example is MMLU (Massive
Multitask Language Understanding) \citep{hendrycks2021mmlu};
it covers 57 diverse fields from basic arithmetic to intricate areas like legal studies and computer science. 
Although many of these benchmarks have focused on English, there have been significant efforts to adapt and develop similar benchmarks for other languages \citep{son2024kmmlu, koto2024arabicmmlu, li2023cmmlu, senel-etal-2024-kardes, conneau-etal-2018-xnli, ponti-etal-2020-xcopa}.

\begin{figure*}[t]
    \centering    \includegraphics[width=0.8\textwidth]{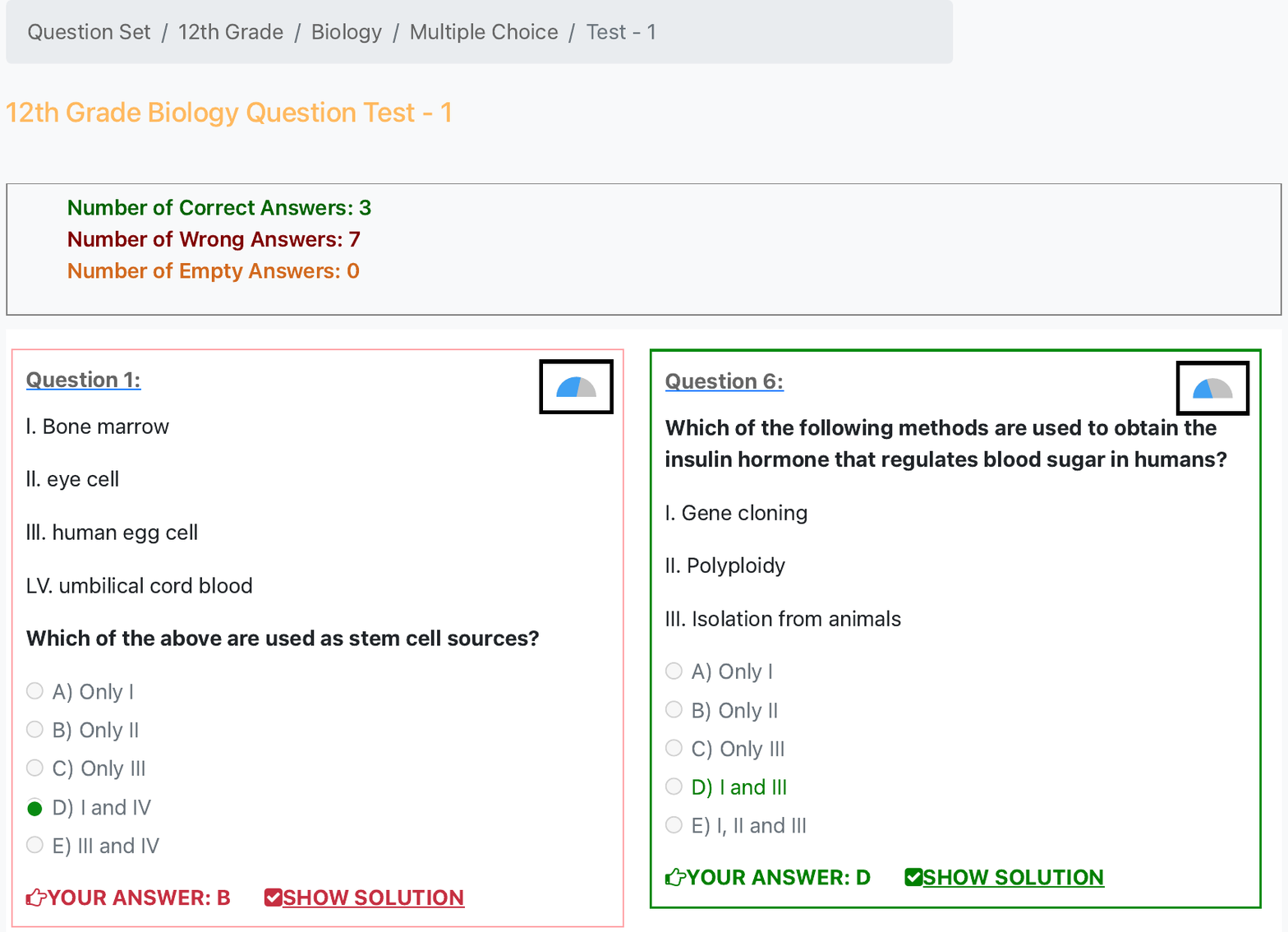}
    \caption{Sample biology test from the EBA Platform (translated to English, see Figure \ref{fig:eba_original} in Appendix for the original Turkish test). 
    Black boxes indicate the correctness ratio (difficulty level). Green borders appear when the user's choice matches the ground-truth answer, while red borders indicate incorrect choices.}
    \label{fig:eba}
\end{figure*}

\noindent\textbf{Turkish Benchmarks:} 
One of the initial efforts in Turkish benchmarking was
THQUAD \citep{Soygazi2021thquad}, a variant of the SQuAD
question-answering
benchmark \citep{rajpurkar-etal-2016-squad} that focuses on
extracting information from historical passages and
answering questions about Ottoman and Islamic history in an
open-book format.
MUKAYESE \citep{safaya-etal-2022-mukayese},
another Turkish benchmark,  was created by combining multiple existing datasets for various tasks.
However, most of the tasks that are included in MUKAYESE,
such as NER (named entity recognition), sentence
segmentation and spellchecking, do not effectively capture the knowledge and the language understanding capabilities of LLMs due to their low level nature. 
Several other studies that created multilingual benchmarks for specific tasks such as XCOPA (Cross-lingual Choice of Plausible Alternatives) \citep{ponti-etal-2020-xcopa} and XNLI (Cross-lingual Natural Language Inference) \citep{conneau-etal-2018-xnli} also include Turkish among several other languages.
A recent study that focuses on Turkish LLMs \citep{acikgoz2024bridging} created the Turkish versions of the TruthfulQA Multiple
Choice (MC) \citep{lin-etal-2022-truthfulqa} and ARC (AI2 Reasoning Challenge) \citep{DBLP:journals/corr/abs-1803-05457} datasets to evaluate Turkish LLMs.
These benchmarks are constructed by machine translating the English versions of the corresponding datasets, which is usually followed by manual verification and editing to ensure good quality.
Overall, despite some efforts to evaluate the capabilities of LLMs for Turkish, Turkish still lacks a high quality and comprehensive evaluation resource that covers multiple domains. 
In this study, we address this by introducing Turkish MMLU.

\section{Dataset}\label{sec:dataset}

\dataset is curated using resources from online learning materials for Turkish high school education. In the Turkish educational system, high school education spans four years, and students take the National University Entrance Exams after completing their studies. 
This exam contains multiple-choice questions covering various subjects from the curricula. 
To assist students in preparing for these exams, official and commercial exam preparation booklets, video guides, 
and online practice tests in multiple-choice question-answering format are available. The Turkish Ministry of Education (MEB) has developed 
an online platform called the Education Information Network (EBA), which aims to provide electronic resources such as lecture notes, videos, 
tests and solutions, and interactive books to facilitate the learning process for students. 
This platform\footnote{\href{https://ogmmateryal.eba.gov.tr/panel/MSoruDers.aspx}{https://ogmmateryal.eba.gov.tr/panel/MSoruDers.aspx}} contains 
multiple-choice questions and their solutions that form the basis of our study.
The questions are driectly sourced from the platform which ensures that each question is crafted and verified by domain experts, guaranteeing correctness 
and high quality.

We conducted repeated manual reviews, randomly selecting 30 questions from the complete dataset during each review
session to assess their formatting and accuracy. Whenever we detected formatting issues—typically due to irregular presentation 
on the source website—we revised our parsing code to effectively address these edge cases. 
This process was continued until we ensured consistent formatting accuracy across multiple consecutive rounds of the sampled questions.

\begin{figure}[htb]
    \centering
    \includegraphics[width=\columnwidth]{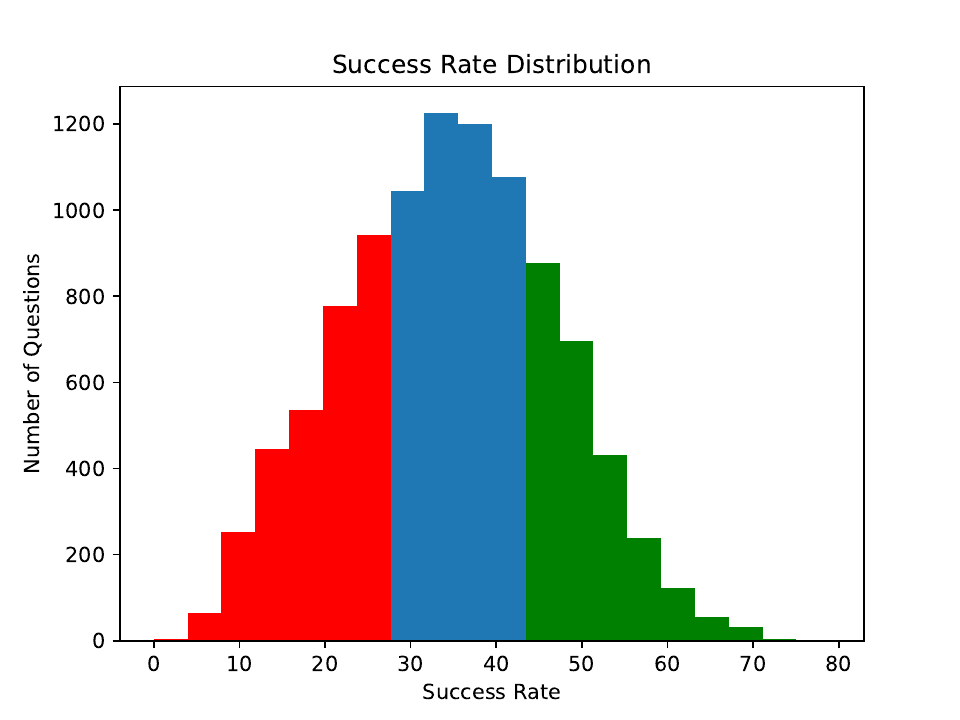}
    \caption{Distribution of correctness ratios. Questions are categorized as Easy (green, top 30\%), Medium (blue, middle 40\%), or Hard (red, bottom 30\%) based on the 30th and 70th percentiles.}
    \label{fig:dataset_success_rate}
\end{figure}

Figure \ref{fig:eba} illustrates the EBA platform interface. Users generate tests by specifying grade level and subject, upon which the platform provides multiple 10-question tests. After test completion, users can review ground-truth answers and video solutions. Each question's difficulty is denoted by a Correctness Ratio (black boxes in Figure \ref{fig:eba}), calculated as the percentage of correct user responses. For each test, we extract question text, multiple-choice options, correct answer, topic, subject, grade, and difficulty level.

Table \ref{tab:dataset_test_subject_details} details the distribution of test questions by grade and subject in \dataset. The dataset includes nine high school subjects across four domains: Math (Mathematics); Natural Sciences (Biology, Chemistry, Physics); Language (Turkish Language and Literature); and Humanities and Social Sciences (History, Geography, Philosophy, Religion and Ethics). The test set comprises 9,807 multiple-choice questions, with an additional 225 (25 per subject) in the development set. While Philosophy is limited to grades 10 and 11, other subjects span all four grades. Many questions include mathematical formulas/notations (in LaTeX or text) and images, however, we exclude image-based questions to focus on evaluating text models.

Figure \ref{fig:dataset_success_rate} displays the distribution of Correctness Ratios. Questions are categorized as Easy (top 30\%), Medium (middle 40\%), or Hard (bottom 30\%), with percentile thresholds at 41 and 28, respectively.

We manually selected 25 questions per
subject for the development set, maintaining subject-grade distributions and mirroring the overall difficulty distribution. For few-shot examples, we focus on 5-shot experiments with 5 questions per subject due to context window constraints and compute budget limitations, each with different correct answers to avoid selection bias. For Chain-of-Thought (COT) prompting, we manually provide step-by-step solutions for these 5 questions per subject.

The large scale of our test dataset, including 9,807
questions, raises significant challenges. Experiments with
state-of-the-art proprietary models like GPT 4 and
Claude-Opus face budget constraints, while using
Chain-of-thought (COT) prompting
with
open-source models
generates excessively long responses, resulting in long
inference times. To address these issues while maintaining
comprehensive evaluations, we create a smaller version
of \dataset, called \datasetsmall with 100 randomly selected questions per
subject, totaling 900. We uniformly sampled 25 questions per
grade for each subject, except for Philosophy, which has 50
questions evenly distributed between grades 10 and 11.
This sample is representative of
grades and subjects, enabling in-depth model evaluation, but
can be easily used in resource-constrained scenarios. We measure the correlation between \datasetsmall and \dataset in \S\ref{sec:correlation_small}, finding a strong correlation across 32 models.

\begin{table}[t]
\centering
\resizebox{\columnwidth}{!}{
\begin{tabular}{lrrrrr}
\toprule
\textbf{Subject} & \multicolumn{4}{c}{\textbf{Grade}} & \textbf{Total} \\ 
& 9 & 10 & 11& 12 & \\
\midrule
Turkish L \& L & 251 & 336 & 208 & 246 & 1041 \\
Mathematics & 565 & 470 & 64 & 379 & 1478 \\
Physics & 194 & 93 & 78 & 246 & 611 \\
Chemistry & 283 & 474 & 340 & 309 & 1406 \\
Biology & 273 & 328 & 401  & 323  & 1325 \\
History & 342 & 398 & 281 & 316 & 1337 \\ 
Geography & 331 & 364 & 494 & 290 & 1479 \\
Religion and Ethics& 120 & 229 & 122 & 42 & 513 \\
Philosophy & 0 & 332 & 285 & 0 & 617\\
\bottomrule
\end{tabular}
}
\caption{Distribution of test questions of \dataset by subject and grade.}\label{tab:dataset_test_subject_details}
\end{table}

\section{Evaluation Results}\label{sec:results}
After finalizing \dataset, we now evaluate various multilingual and Turkish-adapted open- and closed-source LLMs. We cover a wide range of models, from 60M to 141B parameters, and various experimental setups.

\paragraph{Experimental Setup}
Our main evaluation setup is \textbf{5-shot in-context
learning evaluation}, following the prior evaluation setups
in recent LLMs \citep{gemmateam2024gemma, openai2024gpt4} on
English MMLU \citep{hendrycks2021mmlu}.
From the development set proposed in \S\ref{sec:dataset}, we select a fixed
set of questions for each subject and include 5 of them in
our few-shot prompt, with the question, multiple-choice
options, and the answer. We carefully design these prompts
to ensure that each question has a different option
(in our dataset, the five options are always A, B, C, D, E)
as the
answer.
For
evaluation, we report accuracy by using the
lm-evaluation-harness framework from
EleutherAI \citep{eval-harness}.
For open-source models,
we perform log-prob based
evaluation;
for closed-source models we perform
greedy
decoding and then parse the prediction. In order to ensure validity of parsing,
we utilize regex patterns based on the few-shot example structure which can be formulated as \texttt{Question: ...; Choices: A. ... E. ... ; Answer: [A-E] }.

Our second evaluation is a \textbf{zero-shot evaluation} to
 compare few-shot and zero-shot performance of the
 models. Additionally, we evaluate LLMs with
 a \textbf{5-shot chain-of-thought (CoT)
 evaluation}. Especially for questions requiring further
 reasoning and elaboration, such as mathematics, directly
 giving answers may be a limitation in our main
 evaluation. Therefore, we evaluate a wide range of models,
 including closed-source models, with CoT
 reasoning \citep{wei2022chain}. In this setup, we provide
CoT solutions for each question in our few-shots for each
 subject and perform greedy decoding. We put the final
 answer option at the end of the solution in the prompts,
 and then parse the predicted answer in the generated
 solution. We formatted the CoT few-shot examples as they 
 would end with the Turkish phrase of \texttt{The correct choice is [A-E]}.
 This way we can parse the predicted answer easily from the generated solution.

Since \dataset includes real-world data for difficulty, we
also conduct a \textbf{difficulty analysis} to evaluate
models. This expands our evaluation setup from comparing
models on different subjects to varying difficulty
levels. In all of our evaluations, we use a small subset
of \dataset, \datasetsmall, because the closed-source experiments are quite
expensive.\footnote{For example, a 5-shot CoT evaluation
with Claude-3 Opus on the entire dataset would cost more
than \$750.} With public models, we calculate performance
on both \dataset and \datasetsmall to
test our assumption that they would yield similar results.

\paragraph{Language Models:}
We evaluate a diverse range of models, including Turkish-adapted, multilingual open-source and closed-source LLMs.

For Turkish-adapted models, we use Trendyol-LLM 7B, a
Llama-2 model further pretrained on Turkish\footnote{\url{https://huggingface.co/Trendyol/Trendyol-LLM-7b-base-v0.1}}, available in base, chat, and chat-dpo forms on HuggingFace. We also include Kanarya \citep{safaya-etal-2022-mukayese}, a pretrained autoregressive 2B Turkish model.

In the multilingual open-source category, we evaluate models with encoder-decoder architectures such as mT5 \citep{xue-etal-2021-mt5} (from small to xxl), mT0 \citep{muennighoffetal2023mt0} (with the same sizes as mT5), and Cohere's Aya-101 \citep{ustun2024aya}. For autoregressive models, we include Meta's Llama-2 \citep{touvron2023llama2} (7B, 7B-Chat, 13B, 13B-Chat) and Llama-3 (8B, 8B-Instruct, 70B, and 70B-Instruct). From MistralAI, we evaluate Mistral 7B variants \citep{jiang2023mistral}, Mixtral 8x22B, and 8x7B \citep{jiang2024mixtral}. We also include Cohere4AI's Command-R and Aya-23 models \citep{aryabumi2024aya}, Google's Gemma \citep{gemmateam2024gemma} (7B and 2B with their instruction versions), and Microsoft's Phi-3-Mini \citep{abdin2024phi3}.

For multilingual closed-source models, we evaluate OpenAI's
GPT models (3.5, 4-Turbo, and 4o), Anthropic's Claude-3
models (Haiku, Sonnet, and Opus versions), and Google's
Gemini models (pro versions
1.0 and 1.5).

\begin{table*}[htb]
\centering
\resizebox{0.9\textwidth}{!}{%
\begin{tabular}{llccccc}
\toprule
\multirow{2}{*}{\textbf{Model}} & \multirow{2}{*}{\textbf{Source}} & \multirow{2}{*}{\textbf{Average}} & \textbf{Natural} & \multirow{2}{*}{\textbf{Math}} & \textbf{Turkish} & \textbf{Social Sciences} \\ 
 & & & \textbf{Sciences} &  & \textbf{L \& L} & \textbf{and Humanities} \\
\midrule
GPT 4o$^*$ & Closed  & \textbf{83.1} & \textbf{75.3} & \textbf{59.0} & \textbf{82.0} & \textbf{95.3}\\
Claude-3 Opus$^*$&	Closed &	79.1&	71.7 &	\textbf{59.0} &	77.0 &	90.3\\
GPT 4-turbo$^*$ & Closed & 75.7 & 70.3 & 57.0 & 67.0 & 86.5\\
Llama-3 70B-IT & Open& 67.3 & 56.7 & 42.0 & 57.0 & 84.3\\
Claude-3 Sonnet$^*$ & Closed& 67.3 & 67.3 & 44.0 & 58.0 & 75.5\\
Llama-3 70B & Open & 66.1 & 56.0 & 37.0 & 57.0 & 83.3\\
Claude-3 Haiku$^*$ & Closed  & 65.4 & 57.0 & 40.0 & 61.0 & 79.3\\
Gemini 1.0-pro & Closed  & 63.2 & 52.7 & 29.0 & 63.0 & 79.8\\
C4AI Command-r+ & Open & 60.6 & 50.0 & 26.0 & 57.0 & 78.0\\
Aya-23 35B &	Open &	55.6 & 43.3 &	31.0 & 49.0	& 72.5\\
C4AI Command-r & Open & 54.9 & 44.7 & 29.0 & 49.0 & 70.5\\
Mixtral 8x22B & Open & 54.8 & 45.3 & 27.0 & 49.0 & 70.3\\
GPT 3.5-turbo$^*$ & Closed & 51.0 & 42.7 & 39.0 & 45.0 & 61.8\\
Llama-3 8B-IT & Open & 46.4 & 36.7 & 29.0 & 39.0 & 60.0\\
Llama-3 8B & Open  & 46.2 & 37.3 & 30.0 & 33.0 & 60.3\\
Mixtral 8x7B-IT & Open & 45.2 & 41.3 & 28.0 & 39.0 & 54.0\\
Aya-23 8B	& Open & 45.0 & 39.0 & 23.0&	31.0& 58.5\\
Gemma 7B & Open& 43.6 & 34.3 & 22.0 & 47.0 & 55.0\\
Aya-101 & Open & 40.7 & 31.3 & 14.0 & 38.0 & 55.0\\
Trendyol-LLM 7B-C-D  & Open & 34.1 & 30.3 & 22.0 & 28.0 & 41.5\\
mT0-xxl  & Open & 33.9 & 29.3 & 28.0 & 21.0 & 42.0\\
Mistral 7B-IT  & Open & 32.0 & 34.3 & 26.0 & 38.0 & 30.3\\
Llama-2 7B & Open & 22.3 & 25.3 & 26.0 & 20.0 & 19.8\\
mT5-xxl  & Open  & 18.1 & 19.3 & 24.0 & 14.0 & 16.8\\
\bottomrule
\end{tabular}
}
\caption{5-shot experiments on \datasetsmall. Many closed models shift to
chain-of-thought-like detailed explanations, we indicate
this with the $^*$ symbol. Natural Sciences consists of 
Biology, Chemistry, and Physics.
Turkish L\&L is the Turkish Language and
Literature subject. Social Sciences and Humanities consists of History, Geography, Philosophy, and Religion and Ethics.}
\label{tab:5-shot-all}
\end{table*}

\subsection{Few-shot Evaluation}

We present the 5-shot evaluation of models in Table \ref{tab:5-shot-all}. We show scores in four categories: Natural Sciences, Math, Turkish Language \& Literature, and Social Sciences and Humanities, as well as the macro-averaged scores over nine subjects.
The best-performing model is a closed-source model, GPT 4o, with \textbf{83.1\%} accuracy. It outperforms all other models in each category as well. The best-performing open-source model is Llama-3 70B-IT (Instruction-Tuned) with \textbf{67.3\%} accuracy. While it is better than many closed-source models such as Claude-3 Sonnet and Gemini 1.0-pro, it is still 15.8\% worse than GPT 4o. Another interesting point is that the best encoder-decoder model, Aya-101, performs much worse than autoregressive models, achieving only 40.7\% accuracy.

The results suggest that mathematics is the most difficult
subject for almost all models, as it is usually challenging
to answer these questions correctly in a single token, given
that they require multi-hop reasoning. The easiest category
in \datasetsmall is Social Sciences and Humanities. For STEM
courses, models perform poorly compared to other
subjects. We also observe that many closed-source models
switch to COT-like problem-solving rather than providing the
answer directly, even though we provided single-answer style
few-shots. We parse the predicted
option in those answers with manually-designed patterns and
indicate these ``CoT'' models with the * symbol in Table \ref{tab:5-shot-all}.

Among 7B-8B models, Llama-3 8B-IT exhibits the best performance, but Aya-23 and Gemma show comparable results. Mistral 7B-IT and Llama-2 7B lag more than 10\% behind these three models. Among mT5-xxl (13B) based models, Aya-101 achieves the best performance, however, encoder-decoder based models perform worse than autoregressive models of similar sizes.

We note that recent open-source models such as Llama-3, Command-R, Aya-23, and Mixtral 8x22B (all released after April 2024) outperform older closed-source models like GPT 3.5 (released in March 2022), signaling promise for open-source models. However, Turkish-adapted models like Trendyol-LLM, despite outperforming their base model (Llama-2 7B), are significantly behind newer variants of similar size (Llama-3 8B).

We provide the results for all nine subjects and all models in the Appendix in Table \ref{tab:5-shot-appendix}.

\begin{table*}[htb]
    \centering
\resizebox{\textwidth}{!}{    
\begin{tabular}{lll@{\hspace{0.1cm}}rl@{\hspace{0.1cm}}rl@{\hspace{0.1cm}}rl@{\hspace{0.1cm}}rl@{\hspace{0.1cm}}r}
\toprule
\multirow{2}{*}{\textbf{Model}} & \multirow{2}{*}{\textbf{Source}} & \multicolumn{2}{c}{\multirow{2}{*}{\textbf{Average}}} & \multicolumn{2}{c}{\textbf{Natural}} & \multicolumn{2}{c}{\multirow{2}{*}{\textbf{Math}}} & \multicolumn{2}{c}{\textbf{Turkish}} & \multicolumn{2}{c}{\textbf{SocSci/}} \\ 
 & &                    \multicolumn{2}{c}{}
 & \multicolumn{2}{c}{\textbf{Sciences}} &
\multicolumn{2}{c}{}
& \multicolumn{2}{c}{\textbf{L \& L}} & \multicolumn{2}{c}{\textbf{Humanities}} \\
\midrule
GPT 4o & Closed & \textbf{88.2} &(\textcolor{teal}{+5.1}) & \textbf{86.3} &(\textcolor{teal}{+11.0}) & \textbf{84.0} &(\textcolor{teal}{+25.0}) & \textbf{81.0} &(\textcolor{red}{--1.0}) & \textbf{92.5} &(\textcolor{red}{--2.8}) \\
Claude-3 Opus &	Closed & 81.8 &(\textcolor{teal}{+2.7})&	77.0 &(\textcolor{teal}{+5.3})&	74.0 &(\textcolor{teal}{+15.0}) &	76.0 &(\textcolor{red}{--1.0}) &	88.8 &(\textcolor{red}{--1.5})\\

GPT 4-turbo & Closed & 79.2 &(\textcolor{teal}{+3.5}) & 75.3 &(\textcolor{teal}{+5.0}) & 75.0 &(\textcolor{teal}{+18.0}) & 69.0 &(\textcolor{teal}{+2.0}) & 85.8 &(\textcolor{red}{--0.8}) \\
Gemini 1.5-pro* & Closed & 70.1 &(\textcolor{teal}{+45.1}) & 65.0 &(\textcolor{teal}{+43.7}) & 51.0 &(\textcolor{teal}{+27.0}) & 54.0 &(\textcolor{teal}{+7.0}) & 82.7 &(\textcolor{teal}{+60.2}) \\
Llama-3 70B-IT & Open & 68.1 &(\textcolor{teal}{+0.8}) & 62.0 &(\textcolor{teal}{+5.3}) & 57.0 &(\textcolor{teal}{+15.0}) & 53.0 &(\textcolor{red}{--4.0}) & 79.2 &(\textcolor{red}{--5.0}) \\
Claude-3 Haiku & Closed & 66.1 &(\textcolor{teal}{+0.7}) & 56.7 &(\textcolor{red}{--0.3}) & 45.0 &(\textcolor{teal}{+5.0}) & 64.0 &(\textcolor{teal}{+3.0}) & 79.0 &(\textcolor{red}{--0.3}) \\
Llama-3 70B & Open & 63.3 &(\textcolor{red}{--2.8}) & 57.3 &(\textcolor{teal}{+1.3}) & 34.0 &(\textcolor{red}{--3.0}) & 54.0 &(\textcolor{red}{--3.0}) & 77.5 &(\textcolor{red}{--5.8}) \\
Claude-3 Sonnet & Closed & 60.7 &(\textcolor{red}{--6.6}) & 58.7 &(\textcolor{red}{--8.6}) & 38.0 &(\textcolor{red}{--6.0}) & 62.0 &(\textcolor{teal}{+4.0}) & 67.5 &(\textcolor{red}{--8.0}) \\
GPT 3.5-turbo & Closed & 58.2 &(\textcolor{teal}{+7.2}) & 52.3 &(\textcolor{teal}{+9.6}) & 42.0 &(\textcolor{teal}{+3.0}) & 51.0 &(\textcolor{teal}{+6.0}) & 68.5 &(\textcolor{teal}{+6.7}) \\
Gemini 1.0-pro & Closed & 54.1 &(\textcolor{red}{--9.1}) & 42.7 &(\textcolor{red}{--10.0}) & 39.0 &(\textcolor{teal}{+10.0}) & 48.0 &(\textcolor{red}{--15.0}) & 68.0 &(\textcolor{red}{--11.8}) \\
C4AI command-r & Open & 49.6 &(\textcolor{red}{--5.3}) & 40.0 &(\textcolor{red}{--4.7}) & 28.0 &(\textcolor{red}{--1.0}) & 41.0 &(\textcolor{red}{--8.0}) & 64.2 &(\textcolor{red}{--6.2}) \\
Llama-3 8B-IT & Open & 40.6 &(\textcolor{red}{--5.8}) & 35.0 &(\textcolor{red}{--1.7}) & 20.0 &(\textcolor{red}{--9.0}) & 29.0 &(\textcolor{red}{--10.0}) & 52.8 &(\textcolor{red}{--7.2}) \\
Mixtral 8x7B-IT & Open & 40.1 &(\textcolor{red}{--5.1}) & 33.0 &(\textcolor{red}{--8.3}) & 33.0 &(\textcolor{teal}{+5.0}) & 39.0 &(\textcolor{teal}{+0.0}) & 47.5 &(\textcolor{red}{--6.5}) \\
Gemma 7B & Open & 34.0 &(\textcolor{red}{--9.6}) & 26.3 &(\textcolor{red}{--8.0}) & 17.0 &(\textcolor{red}{--5.0}) & 27.0 &(\textcolor{red}{--20.0}) & 45.8 &(\textcolor{red}{--9.2}) \\
Llama-3 8B & Open & 28.2 &(\textcolor{red}{--18.0}) & 24.3 &(\textcolor{red}{--13.0}) & 7.0 &(\textcolor{red}{--23.0}) & 27.0 &(\textcolor{red}{--6.0}) & 36.8 &(\textcolor{red}{--23.5}) \\
Trendyol-LLM 7B-C & Open & 27.7 &(\textcolor{red}{--10.3}) & 24.0 &(\textcolor{red}{--6.3}) & 6.0 &(\textcolor{red}{--12.0}) & 26.0 &(\textcolor{red}{--9.0}) & 36.2 &(\textcolor{red}{--13.2}) \\
\bottomrule

\end{tabular}
}
\caption{5-shot chain-of-thought (CoT) evaluation results on \datasetsmall. The table presents accuracy for four subject categories and the macro-average, with performance changes from non-CoT experiments in parentheses. \\{\small* Gemini 1.5-pro's large improvement (+45.1) is due to a model behavior that causes mispredictions in non-CoT, rather than true CoT gains.}}
\label{tab:cot-all}
\end{table*}

\subsection{Zero-Shot Evaluation}

To assess the performance gain from few-shots, we also compare models in zero-shot settings. Table \ref{tab:zero-shot-all} summarizes the results for selected open-source models. We observe the most significant performance improvement via few-shot in the Gemma 7B model. Llama-3 70B-IT, the best-performing model in the few-shot setting, also leads in the zero-shot setting among public models with a minimal performance drop of just 2.7\%.

Interestingly, mT0-xxl performs considerably better in the zero-shot setting than in the few-shot setting, contrary to the trends in the other models. We attribute this to mT0's \citep{muennighoffetal2023mt0} primary focus on zero-shot adaptation. This finding suggests that mT0's zero-shot performance even surpasses Aya's few-shot performance.

\begin{table}[t]
\centering
\resizebox{\columnwidth}{!}{%
\begin{tabular}{lll@{\hspace{0.1cm}}r}
\toprule
\textbf{Model} & \textbf{Zero-Shot}& \multicolumn{2}{c}{\textbf{5-Shot}}      \\ 
\midrule
Llama-3 70B-IT     & \textbf{64.6} & \textbf{67.3} &(\textcolor{teal}{+2.7}) \\
C4AI Command-r+           & 50.6          & 60.6 &(\textcolor{teal}{+10.0})         \\
Mixtral 8x22B            & 46.8          & 54.8 &(\textcolor{teal}{+8.0})         \\
Aya-23 35B & 45.3 & 55.6 &(\textcolor{teal}{+10.3})\\
mT0-xxl                       & 44.8          & 33.9 &(\textcolor{red}{--10.9})          \\
C4AI Command-r               & 42.4          & 54.9 &(\textcolor{teal}{+12.5})        \\
Llama-3 8B-IT      & 38.3          & 46.4 &(\textcolor{teal}{+8.1})           \\
Aya-101                       & 37.4          & 40.7  &(\textcolor{teal}{+3.3})        \\
Mixtral 8x7B-IT    & 35.8          & 45.2 &(\textcolor{teal}{+9.4})       \\
Trendyol-LLM 7B-C-D & 33.3          & 34.1  &(\textcolor{teal}{+0.8})        \\
Mistral 7B-IT      & 24.6          & 32.0   &(\textcolor{teal}{+7.4})       \\
Gemma 7B & 23.1 & 43.6 &(\textcolor{teal}{+20.5}) \\
\bottomrule
\end{tabular}
}
\caption{5-shot and zero-shot accuracy on \datasetsmall for open-source language models. 
}\label{tab:zero-shot-all}
\end{table}

\subsection{Chain-of-Thought Evaluation}
We evaluate 5-shot chain-of-thought (CoT) in Table \ref{tab:cot-all}, showing the performance difference between non-CoT and CoT few-shot experiments. We include CoT evaluations for three reasons: (i) to evaluate reasoning capabilities of recent LLMs, which show promising results \citep{geminiteam2024gemini}, (ii) some subjects like mathematics require multi-hop reasoning, and (iii) CoT also indicates NLG performance of models in Turkish, complementing our NLU evaluation.

All models performing below 60\% accuracy in the non-CoT few-shot scenario, except GPT 3.5-turbo, show worse performance with CoT reasoning. This suggests these models may have limited generation and reasoning capabilities in Turkish. Across all subjects, the most significant improvement is observed in mathematics, with +25.0\% for the best-performing model, GPT 4o. With this approach, GPT 4o sets the best performance on \datasetsmall at 88.2\% accuracy across all settings. We also observe improvements in Natural Sciences, though not as substantial as in Mathematics. However, for Turkish Language \& Literature and Social Sciences and Humanities, we observe no consistent improvements and even performance drops across models, including strong ones.

One exception to our findings is Gemini 1.5-pro. In our
5-shot non-CoT experiments, we found that Gemini 1.5-pro
generates solutions for all questions in the few-shot, even
when provided with gold answers. This prevents us from
getting predictions for test questions since it exceeds our
maximum generation length (it attempts to generate solutions
for 5 few-shot questions + 1 test question). This causes
mispredictions in many 5-shot non-CoT cases for Gemini
1.5-pro. Therefore, the apparent large improvement (+45.1)
between non-CoT and CoT settings for Gemini 1.5-pro is
misleading. In the CoT setting, we see that Gemini is the fourth-best model overall, placing it in a competitive position.

\begin{table}[t]
\resizebox{\columnwidth}{!}{
\begin{tabular}{llrrr}
\toprule
\multirow{2}{*}{Models} & \multirow{2}{*}{r\textsubscript{pb}} & \multicolumn{3}{c}{Accuracy (\%)} \\
\cline{3-5}
& & Easy & Medium & Hard \\
\midrule
GPT 4o & 0.211$^{***}$ & 96.1 & 88.0 & 80.1 \\
Claude-3 Opus & 0.175$^{***}$ & 89.4 & 81.7 & 73.7 \\
GPT 4-turbo & 0.143$^{***}$ & 86.6 & 79.1 & 71.4 \\
Gemini 1.5-pro & 0.228$^{***}$ & 80.3 & 73.7 & 54.5 \\
Llama-3 70B-IT & 0.193$^{***}$ & 79.2 & 68.3 & 56.0 \\
Claude-3 Haiku & 0.265$^{***}$ & 80.6 & 66.0 & 50.8 \\
Llama-3 70B & 0.287$^{***}$ & 76.1 & 68.3 & 43.2 \\
Claude-3 Sonnet & 0.193$^{***}$ & 68.7 & 64.3 & 47.4 \\
GPT 3.5-turbo & 0.220$^{***}$ & 71.1 & 57.1 & 45.9 \\
Gemini 1.0-pro & 0.175$^{***}$ & 65.8 & 52.6 & 43.6 \\
C4AI Command-r & 0.199$^{***}$ & 60.9 & 50.9 & 36.5 \\
Llama-3 8B-IT & 0.197$^{***}$ & 48.9 & 44.0 & 27.1 \\
Mixtral 8x7B-IT & 0.164$^{***}$ & 48.2 & 42.0 & 29.3 \\
Gemma 7B & 0.130$^{***}$ & 40.8 & 34.6 & 25.9 \\
Llama-3 8B & 0.193$^{***}$ & 36.6 & 28.0 & 19.5 \\
Trendyol-LLM 7B-C & 0.152$^{***}$ & 36.6 & 26.6 & 19.5 \\
\bottomrule
\end{tabular}
}
\caption{Chain-of-thought results in \datasetsmall for selected models with respect to question difficulty. The 'r\textsubscript{pb}' column shows the point-biserial correlation coefficient, indicating the strength and direction of the relationship between model performance and question difficulty. All models show a significant positive correlation (p < 0.001), confirming that model performance decreases as question difficulty increases. Easy, Medium, and Hard labels are based on the 30th and 70th percentiles of the correctness ratio distribution (28\% and 41\%, respectively).}
\label{tab:diff-all}
\end{table}

\subsection{Difficulty Analysis}

We analyze model performance across question difficulty
levels using the correctness ratio in \datasetsmall, categorizing
questions as Easy, Medium, or Hard based on the 30th and
70th percentiles. Table \ref{tab:diff-all} presents these
results along with point-biserial correlation coefficients
(r\textsubscript{pb}), which all show statistically
significant positive correlations (p < 0.001), confirming
that model performance decreases as question difficulty
increases. This pattern holds across all models, from
smaller ones like Trendyol-LLM 7B-C (r\textsubscript{pb} =
0.152) to state-of-the-art models like GPT 4o
(r\textsubscript{pb} = 0.211), \textit{validating the
difficulty categorization in \dataset}. On the other hand,
when we apply point-biserial correlation coefficients to the
grade instead of the question difficulty, we do not observe
any significant correlation (p > 0.1) for any of the models.
Surprisingly, difficult questions at
the lower grades seem to be as hard for models as difficult
questions at the higher grades.
Models generally perform well on easy questions (up to
96.1\% accuracy) but struggle with hard ones (19.5\% to
80.1\%). We also observe that for some models, the largest
differences come from the hard questions. For example,
Gemini 1.5-pro is only 6\% lower than GPT 4-turbo in easy
and medium questions, however the gap is 17\% in hard
questions.

\begin{figure}[t]
    \centering
    \includegraphics[width=0.75\linewidth]{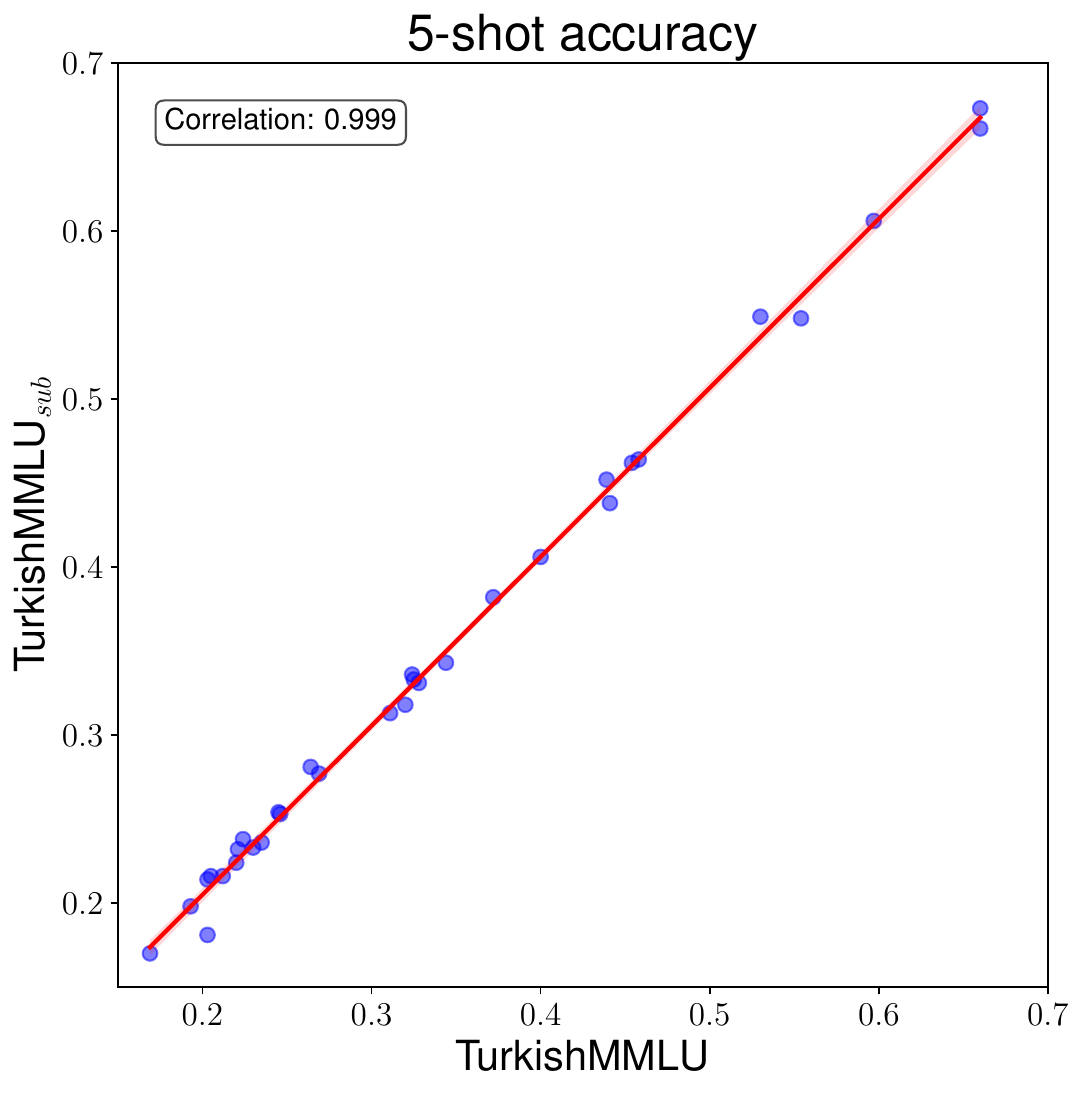}
    \caption{5-shot accuracy comparison of 32 open-source models for \datasetsmall and \dataset (each point corresponds to an LLM). Pearson's r correlation between them is 0.999.}
    \label{fig:small-all}
\end{figure}

\subsection{Small Set - All Set Correlation}
\label{sec:correlation_small}
To reduce the inference time and cost of the experiments,
many analyses in this paper
are conducted on \datasetsmall. In this section, we computed
5-shot average scores for the open-source models in both the
small and full sets. The correlation plot is shown in
Figure \ref{fig:small-all}. Pearson's r correlation between
the two sets is 0.999, confirming that
findings based on \datasetsmall
 are likely to hold as well for \dataset.

\section{Conclusion}\label{sec:conclusion}

In this study, we introduced \dataset, the first Turkish multitask Question Answering benchmark designed for evaluating LLMs. Our dataset consists of 10,032 multiple-choice questions covering nine subjects from the Turkish high school curriculum and university entrance exams, complete with correctness ratios to indicate question difficulty. We evaluated a wide range of LLMs, including Turkish-adapted and multilingual models, in various setups such as zero-shot, few-shot, and chain-of-thought reasoning. Our results highlighted the superior performance of closed-source models like GPT 4o and Claude-3 Opus and the notable improvements in newer open-source autoregressive models like Llama-3 70B-IT. The benchmark demonstrates significant performance variation by subject and question difficulty, emphasizing the strengths and limitations of current LLMs in understanding and reasoning in Turkish. Furthermore, as LLMs
mature, it will become increasingly crucial to shift the 
focus of the field from English to broader coverage 
of the languages of the world. We see \dataset as a promising
contribution towards ensuring that all language communities will
be equally served by NLP in the future.

\section{Limitations}
While we believe \dataset will significantly contribute 
to Turkish NLP and the design of next multilingual LLMs,
it does have some limitations.
First, \dataset is focused solely on text-based assessment. 
Exploring multimodal questions that involve images or audio
is left for future work. Second, the dataset covers high school
curriculum and university entrance exam questions in a
multiple-choice format. However, future efforts should 
aim to expand Turkish benchmarking datasets to include 
assessments of generative abilities and more open-ended questions.
One other limitation of our study is the potential risk of knowledge leakage, 
as some large language models (LLMs) may have been pre-trained on datasets that overlap with or are sourced from similar data used in our benchmarks,
which could artificially inflate their performance. Although the platform does not display gold answers in its interface, models that encountered these questions during pretraining may still perform better than others.
As noted by \citet{pezeshkpour-hruschka-2024-large}, the order of choices may cause variations in model performance; however, we maintained the original order of the choices in our dataset.

\section*{Acknowledgements}
This work was funded by Deutsche Forschungsgemeinschaft
(project SCHU 2246/14-1).

\bibliographystyle{acl_natbib}
\bibliography{anthology,custom}

\clearpage
\appendix

\section{Question Examples}\label{sec:appendix_question}
\begin{figure*}[t]
    \centering    \includegraphics[width=0.8\textwidth]{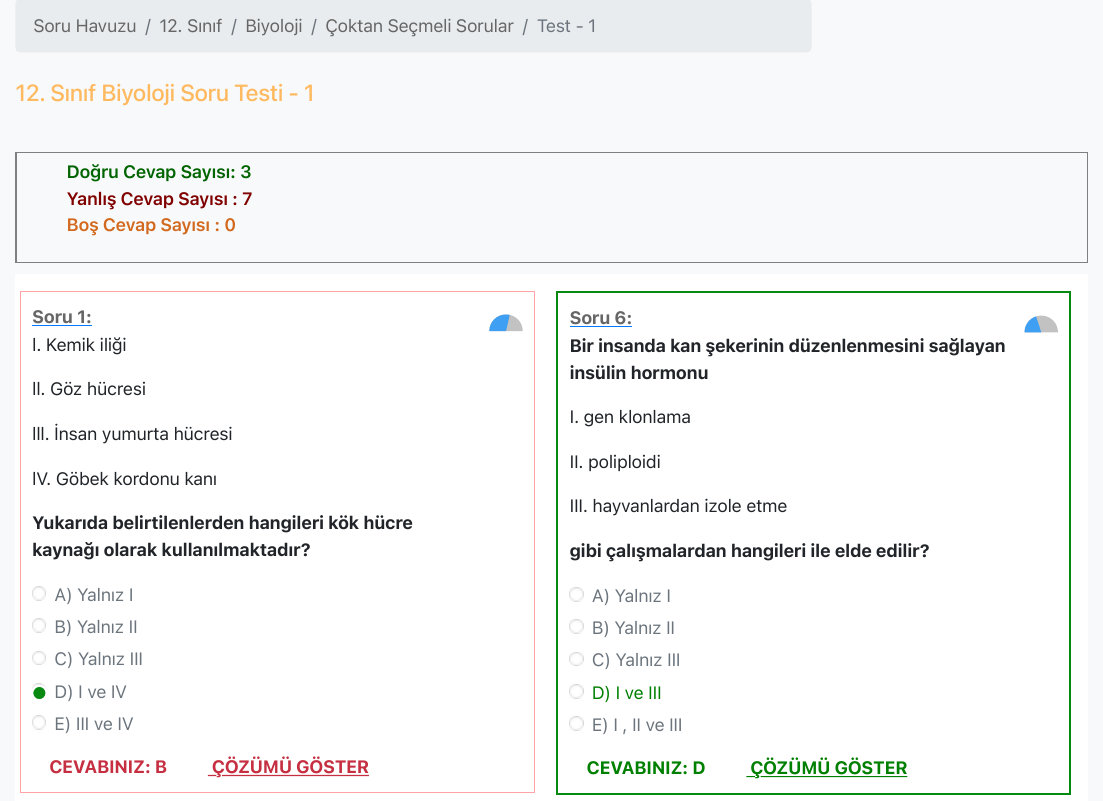}
    \caption{Sample biology test from the EBA Platform, the original version of Figure \ref{fig:eba} in Turkish.} 
    \label{fig:eba_original}
\end{figure*}

\section{Leaderboard}\label{sec:appendix}
For a comprehensive overview of model performance across all nine subjects, we provide detailed leaderboard in this sectoon. Table \ref{tab:5-shot-appendix} presents the 5-shot evaluation scores for 43 models, covering a wide range of LLMs. This detailed breakdown allows for a deeper analysis of model performance variations across different subjects, providing valuable insights into the strengths and weaknesses of each model.

\begin{table*}[t]
\resizebox{\textwidth}{!}{%
\begin{tabular}{llllllllllll}
\hline
\multicolumn{1}{c}{\textbf{Model}} &
  \multicolumn{1}{c}{\textbf{Source}} &
  \multicolumn{1}{c}{\textbf{All}} &
  \multicolumn{1}{c}{\textbf{Biology}} &
  \multicolumn{1}{c}{\textbf{Physics}} &
  \multicolumn{1}{c}{\textbf{Chemistry}} &
  \multicolumn{1}{c}{\textbf{Math}} &
  \multicolumn{1}{c}{\textbf{Turkish}} &
  \multicolumn{1}{c}{\textbf{History}} &
  \multicolumn{1}{c}{\textbf{Geography}} &
  \multicolumn{1}{c}{\textbf{Philosophy}} &
  \multicolumn{1}{c}{\textbf{R\&E}} \\
\hline
GPT 4o &
  Closed &
  \textbf{83.1} &
  78.0 &
  \textbf{77.0} &
  \textbf{71.0} &
  \textbf{59.0} &
  \textbf{82.0} &
  \textbf{96.0} &
  \textbf{95.0} &
  \textbf{98.0} &
  92.0 \\
Claude-3 Opus        & Closed & 79.1 & \textbf{82.0} & 76.0 & 57.0 & 59.0 & 77.0 & 87.0 & 87.0 & 91.0 & \textbf{96.0} \\
GPT 4-Turbo        & Closed & 75.7 & 73.0          & 76.0 & 62.0 & 57.0 & 67.0 & 83.0 & 88.0 & 89.0 & 86.0          \\
Llama-3 70B-IT     & Open   & 67.3 & 59.0          & 59.0 & 52.0 & 42.0 & 57.0 & 86.0 & 85.0 & 85.0 & 81.0          \\
Claude-3 Sonnet     & Closed & 67.3 & 76.0          & 64.0 & 62.0 & 44.0 & 58.0 & 75.0 & 77.0 & 86.0 & 64.0          \\
Llama-3 70B              & Open   & 66.1 & 66.0          & 51.0 & 51.0 & 37.0 & 57.0 & 81.0 & 83.0 & 89.0 & 80.0          \\
Claude-3 Haiku       & Closed & 65.4 & 61.0          & 61.0 & 49.0 & 40.0 & 61.0 & 71.0 & 80.0 & 85.0 & 81.0          \\
Gemini 1.0-pro                & Closed & 63.2 & 63.0          & 53.0 & 42.0 & 29.0 & 63.0 & 76.0 & 75.0 & 86.0 & 82.0          \\
C4AI Command-r+          & Open   & 60.6 & 57.0          & 50.0 & 43.0 & 26.0 & 57.0 & 75.0 & 69.0 & 85.0 & 83.0          \\
Aya-23 35B                    & Open   & 55.6 & 42.0          & 45.0 & 43.0 & 31.0 & 49.0 & 61.0 & 73.0 & 78.0 & 78.0          \\
C4AI command-r           & Open   & 54.9 & 52.0          & 44.0 & 38.0 & 29.0 & 49.0 & 65.0 & 67.0 & 78.0 & 72.0          \\
Mixtral 8x22B           & Open   & 54.8 & 44.0          & 41.0 & 51.0 & 27.0 & 49.0 & 63.0 & 72.0 & 75.0 & 71.0          \\
GPT 3.5-turbo                 & Closed & 51.0 & 47.0          & 43.0 & 38.0 & 39.0 & 45.0 & 58.0 & 57.0 & 72.0 & 60.0          \\
Llama-3 8B-IT      & Open   & 46.4 & 38.0          & 41.0 & 31.0 & 29.0 & 39.0 & 51.0 & 51.0 & 65.0 & 73.0          \\
Llama-3 8B               & Open   & 46.2 & 37.0          & 38.0 & 37.0 & 30.0 & 33.0 & 51.0 & 53.0 & 71.0 & 66.0          \\
Mixtral 8x7B-IT    & Open   & 45.2 & 43.0          & 46.0 & 35.0 & 28.0 & 39.0 & 47.0 & 48.0 & 60.0 & 61.0          \\
Aya-23 8B                     & Open   & 45.0 & 40.0          & 42.0 & 35.0 & 23.0 & 31.0 & 53.0 & 52.0 & 69.0 & 60.0          \\
Gemma 7B                      & Open   & 43.6 & 33.0          & 41.0 & 29.0 & 22.0 & 47.0 & 47.0 & 55.0 & 63.0 & 55.0          \\
Aya-101                       & Open   & 40.7 & 30.0          & 32.0 & 32.0 & 14.0 & 38.0 & 42.0 & 38.0 & 74.0 & 66.0          \\
Trendyol-LLM 7B-C     & Open   & 38.0 & 28.0          & 31.0 & 32.0 & 18.0 & 35.0 & 47.0 & 51.0 & 55.0 & 45.0          \\
Trendyol-LLM 7B-C-D & Open   & 34.1 & 29.0          & 33.0 & 29.0 & 22.0 & 28.0 & 41.0 & 50.0 & 39.0 & 36.0          \\
mT0-xxl                       & Open   & 33.9 & 34.0          & 29.0 & 25.0 & 28.0 & 21.0 & 27.0 & 40.0 & 43.0 & 58.0          \\
Mistral 7B-v0.2            & Open   & 33.1 & 32.0          & 39.0 & 30.0 & 27.0 & 34.0 & 31.0 & 35.0 & 38.0 & 32.0          \\
Mistral 7B-v0.1               & Open   & 32.9 & 31.0          & 39.0 & 26.0 & 28.0 & 31.0 & 29.0 & 35.0 & 43.0 & 34.0          \\
Mistral 7B-IT     & Open   & 32.0 & 32.0          & 39.0 & 32.0 & 26.0 & 38.0 & 20.0 & 35.0 & 40.0 & 26.0          \\
Trendyol-LLM 7B    & Open   & 31.7 & 24.0          & 29.0 & 31.0 & 19.0 & 31.0 & 33.0 & 31.0 & 46.0 & 41.0          \\
mT0-xl                        & Open   & 28.1 & 26.0          & 28.0 & 24.0 & 21.0 & 25.0 & 30.0 & 20.0 & 41.0 & 38.0          \\
Gemma 7B-IT                   & Open   & 27.3 & 28.0          & 26.0 & 26.0 & 25.0 & 25.0 & 25.0 & 30.0 & 31.0 & 30.0          \\
Phi-3-mini-4k-instruct        & Open   & 26.1 & 28.0          & 30.0 & 24.0 & 27.0 & 27.0 & 26.0 & 30.0 & 25.0 & 18.0          \\
Llama-2 13B-C          & Open   & 25.8 & 27.0          & 33.0 & 23.0 & 27.0 & 23.0 & 23.0 & 19.0 & 33.0 & 24.0          \\
Llama-2 13B               & Open   & 25.6 & 28.0          & 28.0 & 24.0 & 31.0 & 22.0 & 23.0 & 25.0 & 25.0 & 24.0          \\
Gemini 1.5-pro                & Closed & 25.0 & 23.0          & 22.0 & 19.0 & 24.0 & 47.0 & 14.0 & 29.0 & 25.0 & 22.0          \\
mT5-base                      & Open   & 23.8 & 26.0          & 25.0 & 19.0 & 19.0 & 21.0 & 30.0 & 28.0 & 23.0 & 23.0          \\
Gemma 2B                      & Open   & 23.4 & 28.0          & 29.0 & 19.0 & 16.0 & 22.0 & 21.0 & 25.0 & 28.0 & 23.0          \\
Gemma 2B-IT                   & Open   & 23.2 & 33.0          & 22.0 & 25.0 & 19.0 & 22.0 & 17.0 & 26.0 & 28.0 & 17.0          \\
Llama-2 7B-C           & Open   & 23.2 & 19.0          & 25.0 & 19.0 & 26.0 & 19.0 & 23.0 & 23.0 & 27.0 & 28.0          \\
Llama-2 7B                 & Open   & 22.3 & 25.0          & 30.0 & 21.0 & 26.0 & 20.0 & 16.0 & 21.0 & 25.0 & 17.0          \\
mT5-xl                        & Open   & 21.6 & 25.0          & 23.0 & 26.0 & 15.0 & 22.0 & 20.0 & 18.0 & 19.0 & 26.0          \\
mT0-large                     & Open   & 21.6 & 16.0          & 16.0 & 27.0 & 23.0 & 21.0 & 19.0 & 19.0 & 26.0 & 27.0          \\
mT0-base                      & Open   & 21.4 & 21.0          & 19.0 & 21.0 & 25.0 & 19.0 & 22.0 & 18.0 & 22.0 & 26.0     \\   
Kanarya 2B                    & Open   & 19.8 & 23.0          & 17.0 & 18.0 & 18.0 & 18.0 & 21.0 & 21.0 & 17.0 & 25.0          \\
mT5-xxl                       & Open   & 18.1 & 19.0          & 20.0 & 19.0 & 24.0 & 14.0 & 19.0 & 19.0 & 17.0 & 12.0          \\
mT5-large                     & Open   & 17.0 & 14.0          & 15.0 & 18.0 & 17.0 & 27.0 & 12.0 & 19.0 & 19.0 & 12.0  \\
\hline
\end{tabular}
}
\caption{5-Shot Experiments for all models on \datasetsmall. The Turkish column refers to the subject of the Turkish Language and Literature, while R\&E is the Religion and Ethics course.}
\label{tab:5-shot-appendix}
\end{table*}

\begin{table*}[t]
    \resizebox{\textwidth}{!}{%
    \begin{tabular}{llllllllllll}
    \hline
    \multicolumn{1}{c}{\textbf{Model}} &
    \multicolumn{1}{c}{\textbf{All (macro)}} &
    \multicolumn{1}{c}{\textbf{All (micro)}} &
    \multicolumn{1}{c}{\textbf{Biology}} &
    \multicolumn{1}{c}{\textbf{Physics}} &
    \multicolumn{1}{c}{\textbf{Chemistry}} &
    \multicolumn{1}{c}{\textbf{Math}} &
    \multicolumn{1}{c}{\textbf{Turkish}} &
    \multicolumn{1}{c}{\textbf{History}} &
    \multicolumn{1}{c}{\textbf{Geography}} &
    \multicolumn{1}{c}{\textbf{Philosophy}} &
    \multicolumn{1}{c}{\textbf{R\&E}} \\
    \hline
    Meta-Llama-3-70B & 66.0 & 63.3 & 65.4 & 59.6 & 53.4 & 30.6 & 58.0 & 77.9 & 78.7 & 89.8 & 81.1 \\
    Meta-Llama-3-70B-Instruct & 66.0 & 63.7 & 62.9 & 62.8 & 53.5 & 35.1 & 57.9 & 76.6 & 80.7 & 87.2 & 77.4 \\
    c4ai-command-r-plus & 59.7 & 56.7 & 51.8 & 48.4 & 42.2 & 25.0 & 56.1 & 74.5 & 73.9 & 87.4 & 77.8 \\
    Mixtral-8x22B-v0.1 & 55.4 & 53.1 & 49.6 & 48.3 & 44.2 & 32.3 & 46.1 & 63.9 & 66.3 & 79.1 & 68.4 \\
    aya-23-35B & 53.9 & 50.6 & 45.6 & 47.3 & 36.9 & 24.2 & 47.6 & 62.6 & 66.6 & 81.7 & 72.3 \\
    c4ai-command-r-v01 & 53.0 & 50.2 & 45.8 & 43.5 & 36.1 & 26.0 & 48.1 & 62.4 & 65.7 & 80.4 & 68.8 \\
    Meta-Llama-3-8B-Instruct & 45.8 & 43.4 & 39.5 & 37.8 & 34.9 & 23.9 & 38.8 & 52.3 & 55.9 & 68.2 & 60.4 \\
    Meta-Llama-3-8B & 45.4 & 43.4 & 39.1 & 33.9 & 36.1 & 25.1 & 39.3 & 51.3 & 55.7 & 72.0 & 56.5 \\
    aya-23-8B & 44.3 & 41.8 & 37.7 & 38.6 & 33.4 & 21.2 & 36.8 & 50.8 & 53.4 & 71.6 & 55.2 \\
    gemma-7b & 44.1 & 41.9 & 36.4 & 37.8 & 34.4 & 25.6 & 36.5 & 47.9 & 54.0 & 69.0 & 55.2 \\
    Mixtral-8x7B-Instruct-v0.1 & 43.9 & 41.9 & 39.7 & 37.6 & 33.1 & 26.9 & 36.9 & 47.7 & 53.4 & 62.7 & 56.7 \\
    aya-101 & 40.0 & 37.2 & 30.6 & 30.1 & 29.3 & 19.7 & 37.8 & 44.8 & 45.0 & 68.4 & 54.2 \\
    Trendyol-LLM-7b-chat-v1.0 & 37.2 & 35.9 & 33.4 & 35.0 & 30.7 & 22.5 & 36.5 & 41.8 & 42.5 & 51.2 & 41.5 \\
    Trendyol-LLM-7b-chat-dpo-v1.0 & 34.4 & 33.4 & 31.2 & 32.1 & 29.2 & 20.5 & 34.2 & 39.1 & 40.3 & 47.6 & 35.7 \\
    Mistral-7B-v0.1 & 32.8 & 32.1 & 32.5 & 31.9 & 31.4 & 23.7 & 28.6 & 34.3 & 36.8 & 43.1 & 32.7 \\
    Mistral-7B-v0.2-hf & 32.5 & 31.9 & 34.3 & 35.7 & 30.7 & 21.2 & 29.1 & 33.6 & 37.6 & 39.5 & 31.0 \\
    mt0-xxl & 32.4 & 30.2 & 30.0 & 34.5 & 26.8 & 20.2 & 19.1 & 28.6 & 38.9 & 42.8 & 50.7 \\
    Mistral-7B-Instruct-v0.2 & 32.0 & 31.4 & 31.1 & 32.4 & 30.4 & 23.5 & 32.5 & 29.7 & 36.8 & 43.1 & 28.3 \\
    Trendyol-LLM-7b-base-v1.0 & 31.1 & 30.0 & 28.7 & 26.2 & 27.2 & 20.2 & 28.3 & 34.7 & 33.9 & 45.9 & 35.3 \\
    gemma-7b-it & 26.9 & 26.4 & 25.4 & 26.2 & 25.7 & 23.3 & 23.7 & 28.9 & 27.2 & 33.4 & 27.9 \\
    mt0-xl & 26.4 & 24.7 & 23.8 & 23.1 & 22.7 & 19.2 & 22.0 & 26.8 & 22.7 & 40.7 & 36.5 \\
    Phi-3-mini-4k-instruct & 25.7 & 25.8 & 26.9 & 28.6 & 27.7 & 24.7 & 23.2 & 25.5 & 25.4 & 30.3 & 18.9 \\
    Llama-2-13b-chat-hf & 24.6 & 24.0 & 23.4 & 31.3 & 23.2 & 20.4 & 22.5 & 26.2 & 22.8 & 32.6 & 19.5 \\
    Llama-2-13b-hf & 24.5 & 24.1 & 25.4 & 25.5 & 23.0 & 21.4 & 22.3 & 24.7 & 24.4 & 29.0 & 24.8 \\
    gemma-2b & 23.5 & 23.2 & 23.4 & 30.6 & 21.9 & 21.4 & 20.8 & 25.1 & 23.1 & 25.3 & 19.7 \\
    gemma-2b-it & 23.0 & 23.0 & 23.2 & 21.3 & 23.0 & 22.3 & 22.9 & 23.8 & 22.6 & 27.9 & 20.1 \\
    mt5-base & 22.4 & 22.1 & 24.0 & 30.1 & 21.9 & 18.6 & 19.1 & 23.1 & 23.1 & 21.1 & 20.5 \\
    Llama-2-7b-chat-hf & 22.1 & 21.9 & 19.4 & 24.4 & 22.9 & 21.4 & 21.2 & 22.5 & 22.3 & 24.3 & 20.3 \\
    Llama-2-7b-hf & 22.0 & 22.0 & 21.3 & 21.8 & 23.6 & 21.4 & 22.2 & 21.7 & 21.4 & 24.3 & 20.3\\
    mt5-xl & 21.2 & 21.1 & 22.0 & 19.0 & 23.1 & 20.1 & 20.7 & 20.7 & 20.2 & 19.9 & 25.1 \\
    mt0-large & 20.4 & 20.2 & 17.7 & 15.9 & 22.7 & 19.9 & 20.6 & 19.9 & 19.1 & 23.0 & 25.1 \\
    mt5-xxl & 20.3 & 20.4 & 22.6 & 21.3 & 21.1 & 20.8 & 20.4 & 19.6 & 19.0 & 20.9 & 17.2 \\
    mt0-base & 20.3 & 20.6 & 18.1 & 17.0 & 23.2 & 21.4 & 20.7 & 20.4 & 21.5 & 21.1 & 18.9 \\
    kanarya-2b & 19.3 & 19.5 & 20.5 & 18.0 & 17.6 & 20.8 & 19.3 & 19.1 & 20.5 & 20.4 & 17.7 \\
    mt5-large & 16.9 & 17.0 & 13.9 & 13.6 & 15.3 & 19.1 & 20.0 & 17.4 & 18.2 & 18.0 & 16.8 \\
    \hline
    \end{tabular}
    }
    \caption{5-Shot Experiments for open models on \dataset. The Turkish column refers to the subject of the Turkish Language and Literature, while R\&E is the Religion and Ethics course.}
    \label{tab:5-shot-all-appendix}
\end{table*}

\end{document}